\documentclass[10pt,journal,compsoc]{IEEEtran}

\ifCLASSOPTIONcompsoc
  \usepackage[nocompress]{cite}
\else
  \usepackage{cite}
\fi

\ifCLASSINFOpdf
  \usepackage[pdftex]{graphicx}
  \graphicspath{{../pdf/}{../jpeg/}}
  \graphicspath{{figures/}{pictures/}{images/}{./}}
  \DeclareGraphicsExtensions{.pdf,.jpeg,.png}
\else
\fi

\usepackage{amsmath}

\usepackage{algorithmic}

\usepackage{array}

\usepackage{url}

\usepackage{amssymb}
\usepackage{multirow}

\usepackage{color}

\newcommand{\ourGAN}{3D-CariGAN}

\newcommand{\yzp}[1]{\textcolor{black}{{#1}}}

\newcommand{\yl}[1]{\textcolor{black}{{#1}}}

\newcommand{\roberto}{\textcolor{black}}
\newcommand{\ylnew}{\textcolor{black}}

\begin{document}
%
\title{3D-CariGAN: An End-to-End Solution to 3D Caricature Generation from Normal Face Photos}
%
%
%
%

\author{Zipeng Ye,
        Mengfei Xia,
        Yanan Sun,
        Ran Yi,
        Minjing Yu,
        Juyong Zhang,~\IEEEmembership{~Member,~IEEE},
        Yu-Kun Lai,~\IEEEmembership{~Member,~IEEE},
        Yong-Jin Liu,~\IEEEmembership{~Senior Member,~IEEE}
\IEEEcompsocitemizethanks{\IEEEcompsocthanksitem 
Z. Ye, M. Xia, Y. Sun, Y.-J. Liu are with BNRist, MOE-Key Laboratory of Pervasive Computing, the Department of Computer Science and Technology, Tsinghua University, Beijing, China.
\IEEEcompsocthanksitem R. Yi is with the Department of Computer Science and Engineering, Shanghai Jiao Tong University, Shanghai, China.
\IEEEcompsocthanksitem M. Yu is with the College of Intelligence and Computing, Tianjin University, China.
\IEEEcompsocthanksitem J. Zhang is with the School of Mathematical Sciences,
University of Science and Technology of China.
\IEEEcompsocthanksitem Y.-K. Lai is with School of Computer Science and
Informatics, Cardiff University, UK
\IEEEcompsocthanksitem
Y.-J. Liu, R. Yi and M. Yu are the corresponding authors. 
}
\thanks{This work was supported by the Natural Science Foundation of China (61725204).}}

%
%

\markboth{Accepted by IEEE Transactions on Visualization and Computer Graphics}%
{Shell \MakeLowercase{\textit{et al.}}: Bare Demo of IEEEtran.cls for Computer Society Journals}
%


\IEEEtitleabstractindextext{%
\begin{abstract}
Caricature is a \roberto{type} of artistic style of human faces that attracts considerable attention in \roberto{the} entertainment industry. So far a few 3D caricature generation methods exist and all of them require some caricature information (e.g., a caricature sketch or 2D caricature) as input. This kind of input, however, is difficult to provide by non-professional users. In this paper, we propose an end-to-end deep neural network model that generates high-quality 3D caricatures directly from a normal 2D face photo. The most challenging issue \roberto{for} our system is that the source domain of face photos (characterized by \roberto{normal 2D} faces) is significantly different from the target domain of 3D caricatures (characterized by 3D exaggerated face shapes and textures).
To address this challenge, we: (1) build a large dataset of 5,343 3D caricature meshes and use it to establish a PCA model in the 3D caricature shape space; (2) reconstruct a \roberto{normal full 3D} head from the input face photo and use its PCA representation in the 3D caricature shape space to \roberto{establish correspondences} between the input photo and 3D caricature shape; and (3) propose a novel character loss and a novel caricature loss based on previous psychological studies on caricatures. 
Experiments including a novel two-level user study show that our system can generate high-quality 3D caricatures directly from normal face photos.
\end{abstract}

}

\maketitle

\IEEEdisplaynontitleabstractindextext
\IEEEpeerreviewmaketitle

\IEEEraisesectionheading{\section{Introduction}\label{sec:introduction}}

\begin{figure}[t]
\small
\centering
\includegraphics[width=\columnwidth]{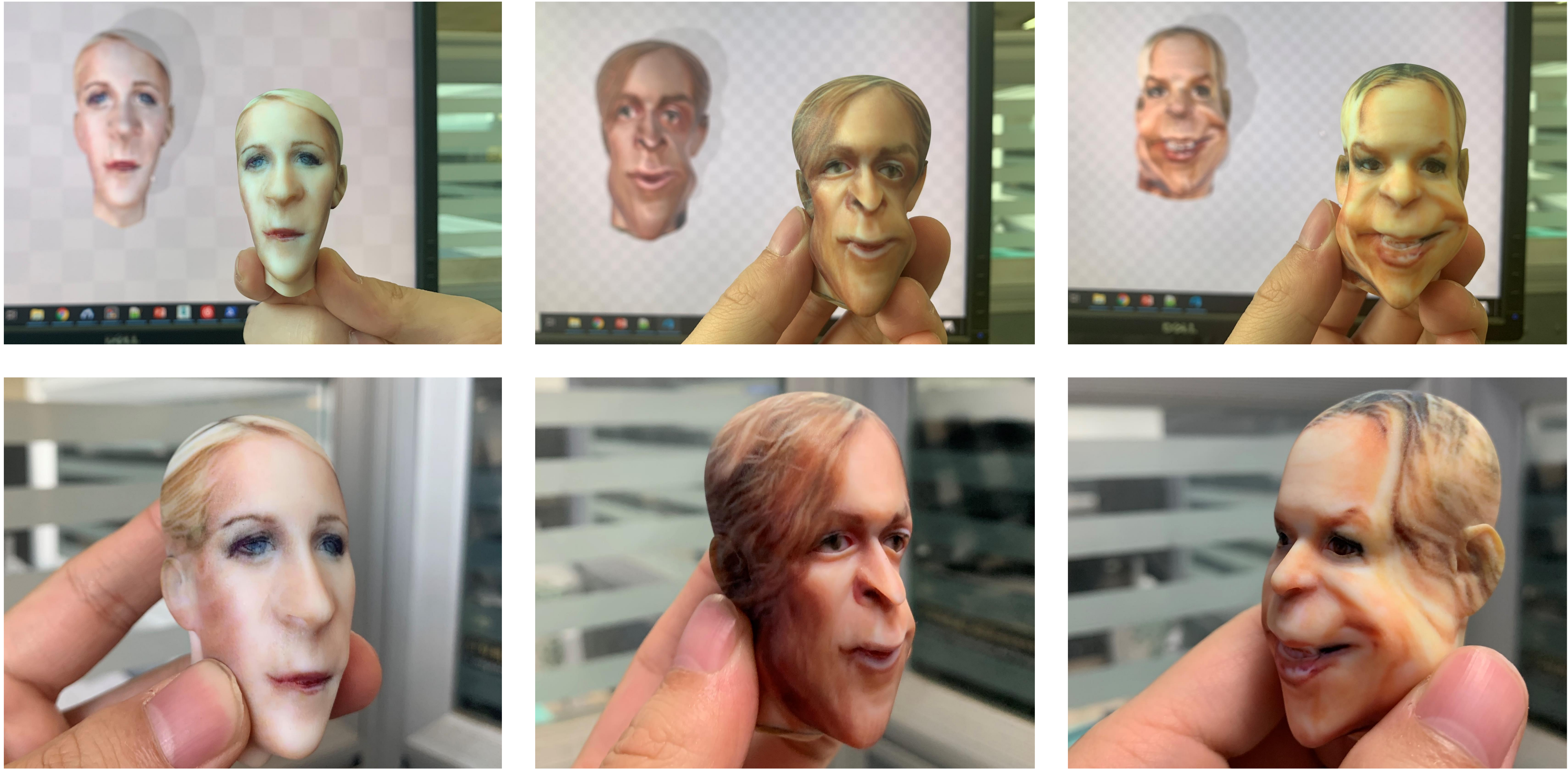}\vspace{-3mm}
\caption{Physical prototypes of 3D caricatures with texture.}
\label{fig:3Dprinting}\vspace{-4mm}
\end{figure}

\IEEEPARstart{T}{radtional} 2D caricature is a \roberto{type} of rendered \roberto{image that uses} exaggeration, simplification and abstraction to express the most distinctive characteristics of people~\cite{sadimon2010computer}. They are also used to express sarcasm and humor for political and social problems.
Traditional caricatures drawn by artists are 2D images. Although widely used, they are insufficient for many graphics applications, such as 3D printing, 3D special effects and animation in feature movies, etc. (Figure~\ref{fig:3Dprinting}). 3D caricatures are \roberto{suitable} for these applications, but it is costly and time consuming \roberto{for artists to create them} with professional 3D modeling \roberto{tools}. 

In this paper, we study the problem of automatic generation of 3D caricatures from {\it normal face photos}. It is an extreme cross-domain task where the input is a normal 2D images and the output is an exaggerated 3D meshes whose space and style are both distinct.
In the literature, efforts have been made to \roberto{address problems that solve part of this task}. Some recent works aim to  translate photos into 2D caricatures~\cite{brennan2007caricature,han2018caricatureshop,li2018carigan,cao2018carigans}. However, image translation mainly focuses on warping and stylization of textures, which only works for 2D images. Wu et al. \cite{wu2018alive} propose a method to reconstruct 3D caricatures from 2D caricatures.  However, its input and output are both caricature styles and the focus is mapping from 2D to 3D.
Han et al.~\cite{han2017deepsketch2face} propose a method for generating 3D caricatures from sketches, but the exaggerated deformation is dominated by sketches, which cannot be applied to normal face photos. Both methods \cite{wu2018alive,han2017deepsketch2face} require some caricature information as input, which is not easy \roberto{for non-professional users to provide}.

In summary, none of \roberto{the} existing methods can automatically generate 3D caricatures directly from normal face photos. For our extreme cross-domain problem, a straightforward baseline approach is to combine several existing methods~\cite{wu2018alive,cao2018carigans,Cole2017synthesizing}, i.e., first automatically generate 2D caricatures from photos and then generate 3D caricatures from 2D caricatures. However, \roberto{this approach} is time-consuming and tends to lose information in intermediate steps --- because the output 3D caricature cannot directly make use of the information in \roberto{the} input photos --- which results in the \roberto{volatility} between the input and the output. In this paper, we present an end-to-end deep learning method for the \roberto{described} problem, which is much more \roberto{efficient and suitably \ylnew{regularized}} by making use of information in input photos as much as possible and allows intuitive control.

Another related \roberto{area} to ours is 3D face reconstruction from photos, which has been widely studied in recent years.  It is popular to use parametric models to represent faces/heads due to the regular structure of faces. Some parametric models such as the well-known 3D Morphable Model (3DMM)~\cite{blanz1999morphable} are linear models based on principal component analysis (PCA)  of normal 3D faces, which are useful and effective. However, such models do not work for caricature faces due to their limited capability of extrapolation~\cite{wu2018alive}.
In this paper, we build a PCA model for 3D caricature meshes, and \roberto{generate} 3D caricature \roberto{models that can} be regarded as interpolation in our PCA space, making the problem more tractable.

Training datasets are indispensable for learning to transform photos to 3D caricatures. For the domain of photos, we use CelebAMask-HQ dataset~\cite{CelebAMask-HQ} which contains \mbox{30,000} portrait photos. For the domain of 3D caricatures, we are not aware of \roberto{any} existing large-scale 3D caricature datasets, so we create our own 3DCari dataset which contains 5,343 3D caricature meshes with the same connectivity. The two datasets are unpaired because it is difficult to obtain the corresponding 3D caricature for a photo.
In this paper, we present an end-to-end method named \ourGAN{} for \roberto{the} automatic generation of 3D caricatures from photos. To train \ourGAN{} using unpaired training data, we propose a novel character loss and a novel caricature loss, both of which are based on previous psychological studies on caricatures \cite{hill2019deep,Benson1991,Rhodes1997}. \ourGAN{} achieves real-time performance and allows users to interactively adjust the caricature facial shapes, \roberto{with simple and effective user controls}. 
Experiments including a novel two-level user study shows that our method produces high-quality 3D caricatures from 2D photos in real-time, which is significantly faster and \roberto{of} better quality than the baseline method. 

In particular, the contributions of this paper include\yzp{\footnote{The dataset, PCA model and source code are available on https://github.com/qq775193759/3D-CariGAN}}:
\begin{itemize}
\item We create a large dataset of 3D caricatures, and based on this, build a novel PCA-based 3D linear morphable model for 3D caricature shapes.
\item We propose the first method to automatically generate 3D caricatures {\it directly} from normal face photos. Our end-to-end solution addresses cross-domain and cross-style challenges (2D to 3D, and normal photo to caricature) by utilizing \roberto{a} caricature morphable model and introducing novel cross-domain character loss and caricature loss.

\end{itemize}

\section{Related Work}

{\bf 2D Caricature.}
Many works have studied generating 2D caricatures from photos. The main differences between photos and caricatures are \roberto{the} 2D geometric shape and image style. Some methods~\cite{brennan2007caricature,han2018caricatureshop} focus on geometric exaggeration \roberto{while} other works~\cite{li2018carigan,gatys2015neural,liao2017visual}
focus on stylization. The work CariGANs~\cite{cao2018carigans} proposes a method that combines these two aspects to generate 2D caricatures using two networks: CariGeoGAN for geometric exaggeration and CariStyGAN for stylization. CariStyGAN disentangles a photo into the style component and the content component, and then replaces the style component by that of a reference or a sample.
CariGeoGAN translates the facial landmarks of a photo from a normal shape to those of an exaggerated shape, which are used to warp the image. WarpGAN~\cite{shi2019warpgan}  generates caricatures by warping and stylization. It extracts the content component from the photo, takes a sample in the style latent space, and then transfers the style by combining the content component and sampled style component, which is similar to CariStyGAN. It warps a photo into a caricature while preserving its identity by predicting a set of control points. The stylization  in these methods can be adapted to stylize textures for 3D caricatures, but geometric exaggeration for 3D caricatures is more complicated, which is a major focus of our paper.

{\bf 3D Face Reconstruction.}
Generating normal 3D faces from photos is well studied in computer graphics. 
The reader is referred to \cite{zollhofer2018state} for a comprehensive survey and \roberto{the} references therein. Due to the regular structure of faces, it is popular to use parametric models to represent faces/heads. 3DMM~\cite{blanz1999morphable,paysan20093d,booth2018large,dai20173d} and multi-linear models~\cite{vlasic2005face,cao2013facewarehouse} are two major types of parametric models. 3DMM is a PCA representation of faces including shapes and textures, and multi-linear models utilize a multi-linear tensor decomposition on attributes such as identity and expression. Parametric models provide a strong constraint to ensure the plausibility of reconstructed 3D face shapes, while substantially reducing the dimensionality of the generation \roberto{space by regressing the parameters. For this reason, they are widely used for face reconstruction}.
Recent works~\cite{jackson2017large,tewari2017mofa,jiang20183d} use convolutional neural networks (CNNs) to regress the parameters for face reconstruction. However, these methods mainly work for normal photos and generate normal 3D faces.
Likewise, existing parametric models do not have enough extrapolation capability to represent 3D caricature faces~\cite{wu2018alive}.  This motivates us to build a new 3D caricature parametric model and a new neural network for unpaired cross-domain translation.

{\bf 3D Caricatures.}
Although generating 3D caricatures from 2D caricatures or normal photos is similar to 3D face reconstruction, only a few works tackle the problem of automatically generating caricatures.
Sela et al.~\cite{sela2015computational} present a method for directly exaggerating 3D face models, which locally amplifies the area of a given 3D face model based on Gaussian curvature.
A deep learning based sketching system~\cite{han2017deepsketch2face} is proposed for interactive modeling of 3D caricature faces by drawing facial contours. A method by Clarke et al.~\cite{clarke2010automatic} generates a 3D caricature from a facial photograph and a corresponding 2D hand-drawn caricature which captures the artistic deformation style.  However, the method requires paired data as input which is difficult to obtain.
An optimization-based method~\cite{wu2018alive} is proposed to reconstruct 3D caricatures from 2D caricatures. This method formulates 3D caricatures as deformed 3D faces. To support exaggeration, their method uses an intrinsic deformation representation which \roberto{is capable} of extrapolation. Therefore, 3D caricature reconstruction is turned into an optimization problem with facial landmark constraints. However, all of these methods rely on 2D sketches or 2D caricatures which contain the information of how to exaggerate and deform \roberto{the 3D surface}, but normal 2D face photos do not have such information. Our work addresses a new challenge of automatically transforming normal 2D face photos to 3D caricatures, without any caricature information used as input.

\begin{figure}[ht]
  \centering
  \includegraphics[width=0.5\textwidth]{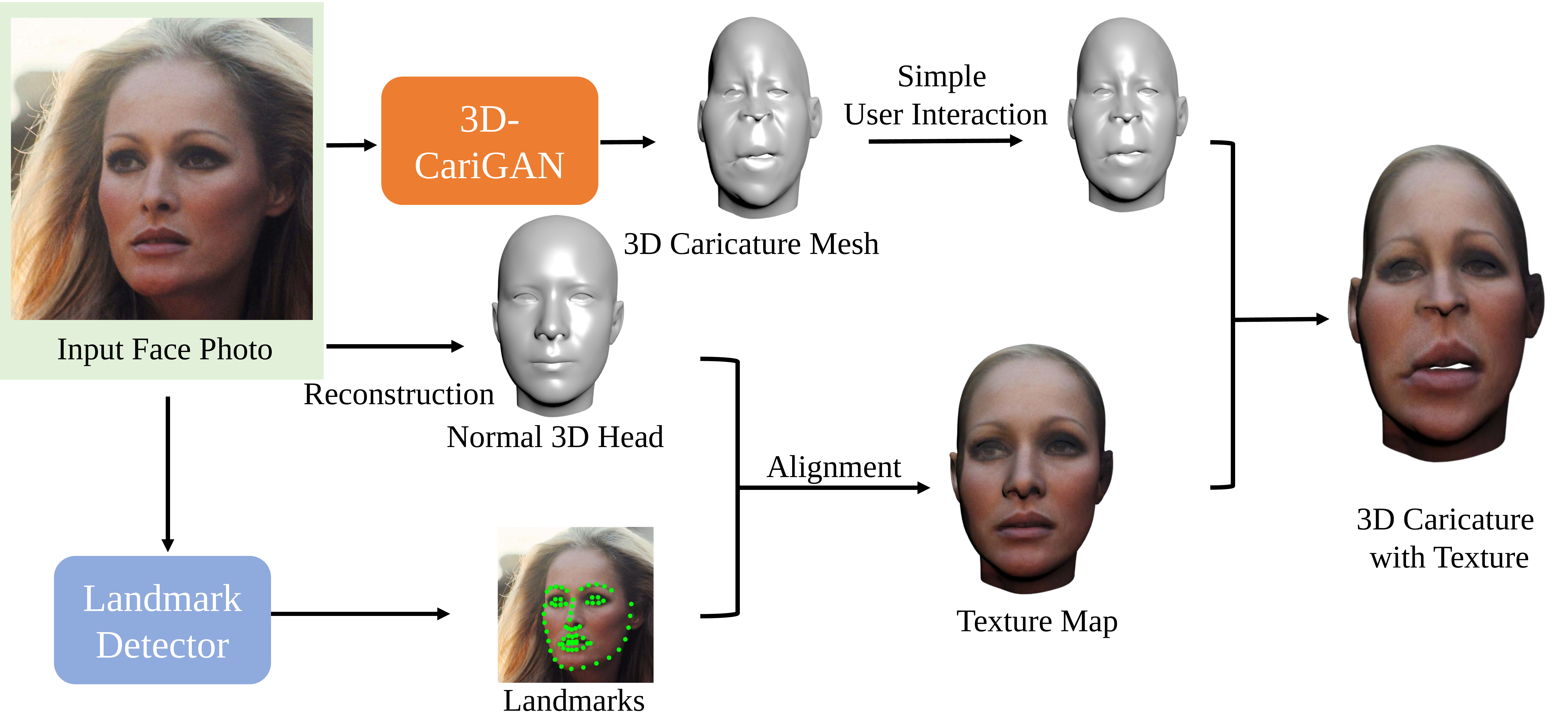}
   \vspace{-0.25in}
  \caption{The pipeline of our method, including \ourGAN{} for transforming photos to 3D caricature meshes, 
  and landmark based texture mapping for generating textured 3D caricatures. Our system also supports a simple and intuitive user interaction for 3D caricature shapes.
  }
  \label{fig:pipeline}
  \vspace{-0.05in}
\end{figure}

\section{Method}


The pipeline of our method is illustrated in Figure~\ref{fig:pipeline}. A 3D caricature consists of three components: a 3D mesh, a 2D texture image and a texture mapping (represented by texture coordinates). Automatically generating 3D caricatures from photos is decoupled into three steps: (1) we develop \ourGAN{} to infer 3D caricature meshes from photos; (2) to map the texture to 3D caricature mesh, we reconstruct a normal 3D head model with the same mesh connectivity as the 3D caricature mesh and consistent landmark positions when projected onto the photo; and (3) we use the input photo as the texture image and the projection matrix \roberto{for} texture mapping (by transferring texture coordinates directly from the normal 3D face to the generated 3D caricature mesh).

\subsection{Representation of 3D Caricatures}
\label{subsec:pca_3d_caricature}

\begin{figure}[th]
  \centering
  \includegraphics[width=.9\columnwidth]{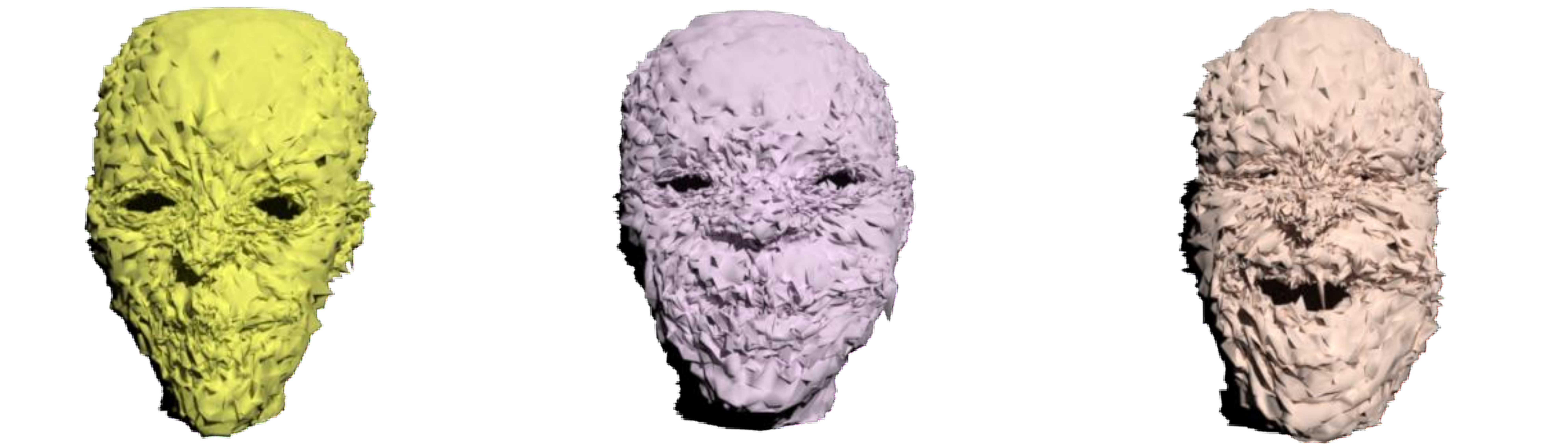}
  \vspace{-4mm}
  \caption{
  3D caricatures obtained by changing our pipeline to directly generate mesh vertex coordinates rather than 3DCariPCA vectors. This alternative \roberto{approach} produces noisy output meshes, due to the higher dimensional space and lack of constraint.
  }\vspace{-5mm}
  \label{fig:noise_mesh}
\end{figure}

Usually a head mesh has tens of thousands of vertices, e.g., 11,510 vertices used in this paper.
A straightforward way is to directly use a neural network for predicting the coordinates of each mesh vertex~\cite{ranjan2018generating}. However, this requires a very large training set and may produce noisy meshes due to insufficient constraints, especially in our cross-domain scenario. Three examples are shown in Figure~\ref{fig:noise_mesh} where we adapt our pipeline to directly generate mesh vertices of 3D caricatures (see Section \ref{subsec:ablation} for more details). The resulting meshes are rather noisy, indicating insufficient constraints due to the amount of training data \roberto{available}. 

\begin{figure}[t]
  \centering
  \includegraphics[width=1.0\columnwidth]{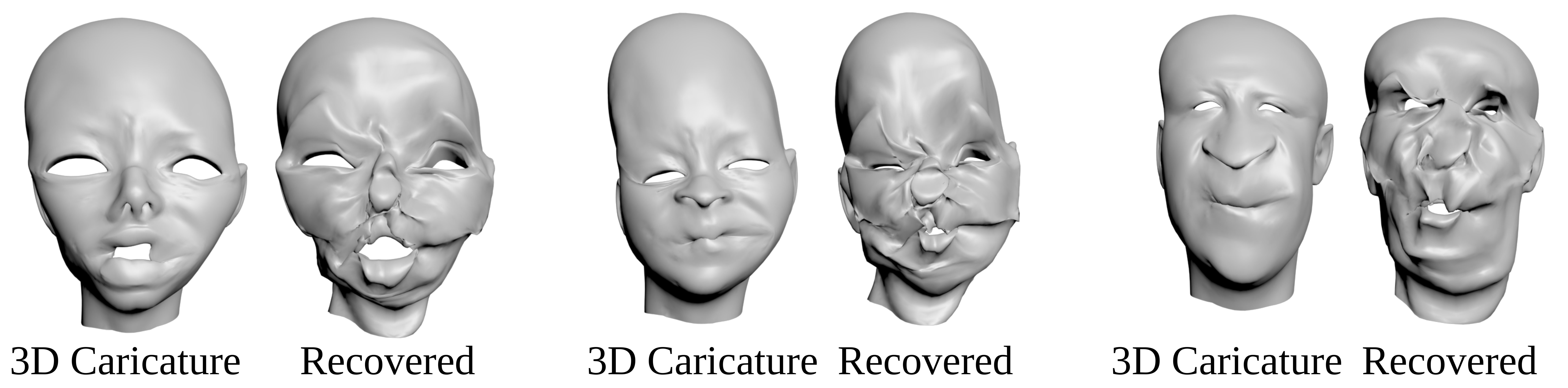}
  \vspace{-7mm}
  \caption{Using the PCA representation of normal faces to represent 3D caricatures. For each group, the left shows the input 3D caricature model and the right shows its corresponding model represented by the normal PCA representation. These examples show that the representation does not have sufficient extrapolation capability to faithfully reconstruct 3D caricatures. This motivates us to create a PCA model for 3D caricatures.
  }\vspace{-5mm}
  \label{fig:pca_extrplotion}
\end{figure}

In normal 3D face reconstruction, to address similar issues, PCA models such as 3DMM \roberto{\ylnew{provide} a strong prior and reduce the dimensionality of the generation space}. However, all the existing PCA models are only \roberto{suitable} for \emph{interpolation} in the shape space of normal 3D faces and they do not work well for \emph{extrapolation} in 3D caricature shape space. To show this, we use a $100$-dimensional PCA representation of normal 3D faces from FaceWarehouse~\cite{cao2013facewarehouse} to represent caricature faces, but the recovered caricatures have substantial distortions, as illustrated in Figure~\ref{fig:pca_extrplotion}. Therefore it is necessary to build a parametric model specifically for 3D caricature shape space.

\begin{figure*}[t]
  \centering
  \includegraphics[width=0.9\textwidth]{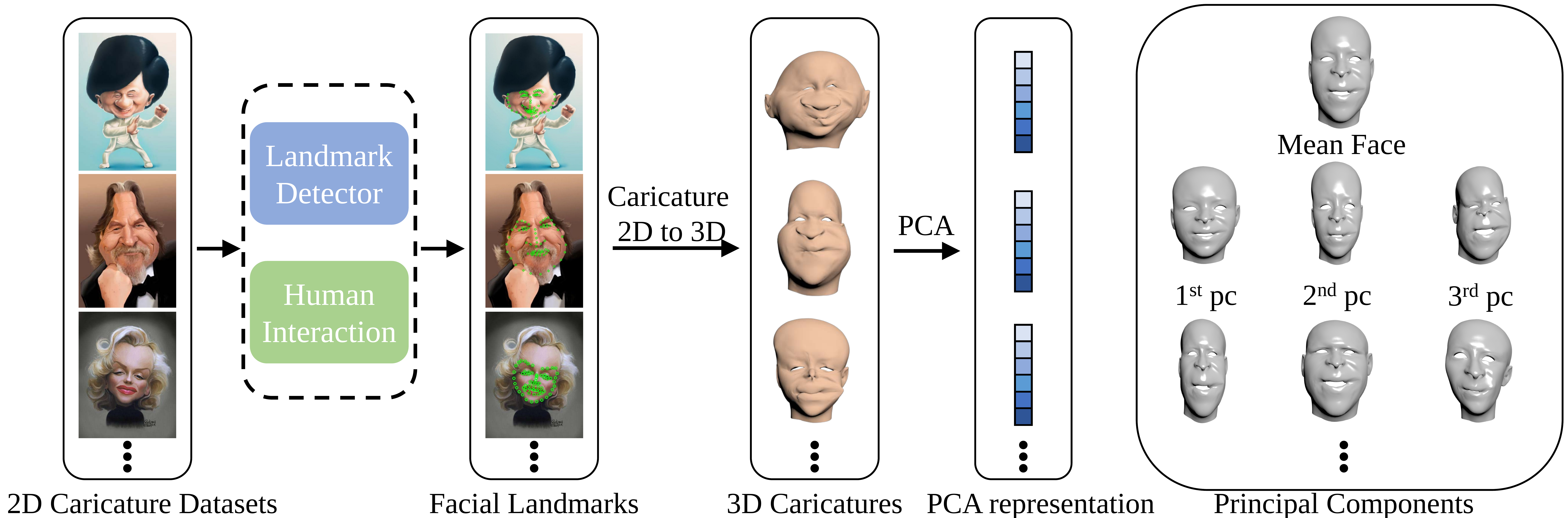}
  \vspace{-0.1in}
  \caption{An illustration of PCA-based 3D caricature representation.  To build this representation, we have collected 5,343 3D caricature meshes by detecting facial landmarks on 2D caricature photos (with manual correction if needed) and reconstructing 3D caricature meshes from them.
 We represent them as a $200$-dimensional vector by applying PCA.
  }
  \label{fig:PCArepresentation}
  \vspace{-0.15in}
\end{figure*}

In our study, we build a PCA model for 3D caricature meshes in the following steps, as illustrated in Figure~\ref{fig:PCArepresentation}.
Since we are not aware of any existing large 3D caricature datasets, we first collect 5,343 hand-drawn portrait caricature images from Pinterest.com and WebCaricature dataset~\cite{Huo:2017:VRC:3126686.3126736,HuoBMVC2018WebCaricature} with facial landmarks extracted by a landmark detector~\cite{king2009dlib}, followed by human interaction for correction if needed. An optimization-based method~\cite{wu2018alive} is then used to generate 3D caricature meshes using facial landmarks of 2D caricatures. We use this method to generate $N$ = 5,343 3D caricature meshes of the same topology. We align the pose of the generated 3D caricature meshes with the pose of a template 3D head using an ICP method, where we use $5$ key landmarks \roberto{on the} eyes, nose and mouth as the \roberto{key point constraints}. We normalize the coordinates of the 3D caricature mesh vertices by translating the center of meshes to the origin and scaling them to the same size. A 3D caricature mesh, denoted as $M_i$ ($1 \leq i \leq N$), can be represented as a long vector containing the coordinates of all the vertices. We apply PCA to all the caricature meshes $\{M_i\}$, 
and obtain the mean head $\bar{h}$ (which is a $3n_v$-vector containing the coordinates of $n_v$ mesh vertices), and $d$ dominant components $\alpha_i$ ($1 \leq i \leq d$). Then, 
a new caricature mesh can be represented as
\begin{equation}
h = \bar{h} + \sum_{i = 1}^{d} h_i \alpha_i = \bar{h} + \mathbf{H}\boldsymbol{\alpha},
\label{eq:linear_combine}
\end{equation}
where $\boldsymbol{\alpha}=(\alpha_1, \alpha_2, \dots, \alpha_{d})^T$ is a collection of $d$ components, and $\mathbf{H} = [h_1, h_2, \dots, h_d]$ is a $d$-dimensional vector that compactly represents the 3D caricature.
The PCA model has only one hyperparameter, i.e, the number of components. Balancing the amount of information against the number of components, we set $d=200$, where the sum of explained variance ratios is $0.9997$, which means it contains almost all the information from the 5,343 meshes. We refer to the above representation as 3D Caricature PCA (3DCariPCA), and use it in our translation network.

In Eq.~(\ref{eq:linear_combine}), the representation is a linear combination of principal components and only involves matrix multiplication, which can be efficiently implemented on \roberto{the} GPU. The gradients of these operations are also linear so the losses related to the mesh can be computed and their gradients can be back-propagated. Therefore, the PCA representation is ideal for our neural-network-based method.

\subsection{Normal Head Mesh Reconstruction from Photos}
\label{subsec:reconstruction_full_head}

In our pipeline, to generate the 3D caricature, we reconstruct a normal 3D head mesh (which has the same connectivity as our caricature meshes) from the input photo. To do so, we first use the method \cite{deng2019accurate}\footnote{We use the source code and the pre-trained model at https:// github.com/changhongjian/Deep3DFaceReconstruction-pytorch} to reconstruct a normal 3D face mesh (i.e., only a front face mesh without ears, neck \roberto{or} the back of the head) from the input photo. Then we use the NICP method~\cite{amberg2007} to register a template 3D head mesh (which is taken from FaceWarehouse~\cite{cao2013facewarehouse}) to the resulting face mesh, and \roberto{simultaneously} use the method~\cite{ploumpis2019combining}\footnote{https://github.com/nabeel3133/combining3Dmorphablemodels} to register the face PCA to the head PCA, 
i.e., building a correspondence between the two PCA models. 
\roberto{The template head mesh registration is performed as follows:} (1) we use facial landmarks as the landmarks for NICP and (2) in the outer loop of NICP, we decrease the stiffness from $50$ to $0.2$ and decrease the landmark weights from $5$ to $0$. 

The registered head model (including ears, neck and the \yl{back of the head}) is used as a bridge to transfer the information from the input photo to the output 3D caricature in two \roberto{parts}: (1) we define a perceptual contrast between a normal 3D head mesh and 3D caricature, to measure the character similarity and caricature style (Section \ref{subsec:generating_caricature}), and (2) we use the texture mapping on \roberto{the normal 3D} head model as the texture mapping on the 3D caricature (Section \ref{subsec:texture}).

\subsection{Generating Caricature Meshes from Photos}
\label{subsec:generating_caricature}

We now describe our network architecture \roberto{for translation from a 2D photo to a 3D caricature}. It is an extreme cross-domain task where the input is \roberto{a normal face image} and the output is \roberto{an exaggerated 3D mesh} whose forms and styles are both totally distinct. We use CelebAMask-HQ dataset \cite{CelebAMask-HQ} which contains 30,000 portrait photos and our 3DCari dataset containing 5,343 full-head 3D caricatures as the training datasets, which are naturally unpaired.

Our network tries to learn the PCA parameters from the photos using a GAN (generative adversarial network) structure, so that it can generate 3D caricature meshes automatically. Denote by $\mathcal{P}$ the domain of photos and $\mathcal{C}$ 
the domain of PCA representation of 3D caricature meshes. The input to our network is a normal face photo $p \in \mathcal{P}$ and the output is a 3D caricature mesh $c \in \mathcal{C}$, represented by our 3DCariPCA representation, to make learning more efficient and incorporate 3D caricature constraint \roberto{to improve} generation results. Our network consists of a generator $G$ and a discriminator $D$. $G$ generates a 3DCariPCA representation $G(p)$ from an input photo $p$, while $D$ discriminates whether an input 3DCariPCA representation is real or synthesized.

\textbf{Network architecture.}
Since our network deals with both 2D images (for which CNNs and residual blocks \cite{he2016deep} are effective) and 3D caricature meshes in the 3DCariPCA space as a $d$-dimensional vector (for which fully connected layers are suitable), both structures are used in our \roberto{architecture}. \roberto{The network} maps a 2D face photo to a 3D caricature mesh. \roberto{\ylnew{This}} involves a sequence of down-sampling convolutional layers, residual blocks, reshaping and fully connected layers. We use 1D batch normalization for fully connected layers and 2D batch normalization for convolutional layers and residual blocks. 

\textbf{Perception of caricatures.} A good caricature optimally selects and exaggerates the most representative facial regions in an artistic way, while humans can still perceive the same identity of the input normal face and the output caricature. To offer a good measure of caricature perception and define good loss terms, we follow the recent neuroscience and psychology studies on caricatures \cite{hill2019deep,Benson1991,Rhodes1997}. These studies show that in a face space $\mathcal{F}$, the difference between a face $f\in\mathcal{F}$ and the neutral face  (i.e., the mean face in 3D normal face space) represent the identity of $f$, and we use this difference as the feature vector of $f$. Then the face identity consistency of two faces in $\mathcal{F}$ can be defined as the cosine of the angle between the feature vectors of them, i.e., \roberto{the} dot product of two normalized feature vectors, we call \roberto{this the} {\it cosine measure}. A good caricature increases the perceptual contrast while maintaining the face identity. In our application, we define an exaggeration equation:
\begin{figure}[t]
\small
\centering
\includegraphics[width=\columnwidth]{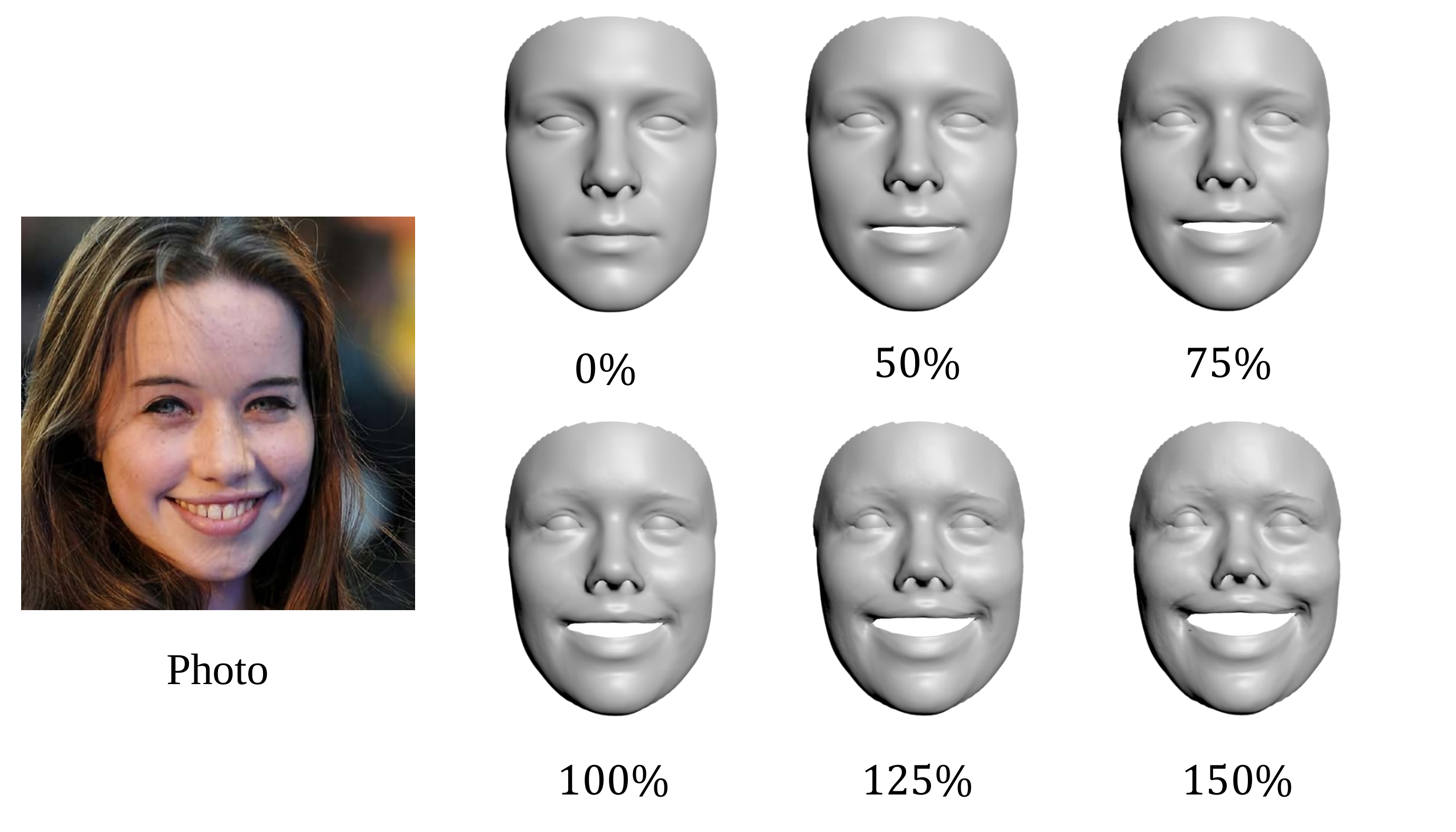}
\vspace{-8mm}
\caption{Exaggeration between a reconstructed face (100\%) and the mean face (0\%). The larger the exaggeration coefficient (125\% and 150\%), the more exaggerated the style is. 
}
\label{fig:perception}\vspace{-5mm}
\end{figure}
\begin{equation}
    C(p, u) = H_{\text{mean}} + u(H(p)-H_{\text{mean}}),
    \label{eq:exaggeration}
\end{equation}
where $H_{\text{mean}}$ is a vector containing coordinates of mean mesh vertices in the normal head space, $u$ is the exaggeration coefficient and $H(p)$ is the reconstructed normal 3D head from the input photo $p$. Note that (1) $H(p)-H_{\text{mean}}$ is the feature vector of $H(p)$; and (2) $H_{\text{mean}}$ is obtained by registering a template head mesh to the mean normal face in BFM model \cite{blanz1999morphable}, which has the same connectivity as caricature meshes. 
All the caricature head models $C(p,u)$ have the same face identity as $H(p)$, and the larger the value $u$ (\roberto{s.t.} $u>1$), the more exaggerated the style is. An example is shown in Figure~\ref{fig:perception}. Below we use the cosine measure to define two novel loss terms.

\textbf{Loss terms.} Adversarial loss $L_{\text{adv}}$ is useful, however, it is insufficient to ensure that the input and the output are the same character, leading to the generation of random persons. We introduce two novel perceptual losses, i.e., character loss $L_{\text{cha}}$ and caricature loss $L_{\text{cari}}$, to constrain the identity of generated caricatures. All of three loss terms $L_{\text{adv}}$, $L_{\text{cha}}$ and $L_{\text{cari}}$ are used for training \ourGAN{}.

{\it Adversarial loss.} $L_{\text{adv}}$ is the adversarial loss which ensures the distribution of $c'=G(p)$ is the same as that of $c$. 
We adapt the adversarial loss of LSGAN~\cite{mao2017least} as
\begin{equation}
\begin{split}
L_{\text{adv}}(G, D, \mathcal{P}, \mathcal{C}) &= \mathbb{E}_{p \sim \mathcal{P}}(\lVert D(G(p)) \rVert_2^2) \\
&+ \mathbb{E}_{c \sim \mathcal{C}}(\lVert 1 - D(c) \rVert_2^2).
\end{split}
\end{equation}
{\it Character loss.} We propose $L_{\text{cha}}$ which aims to measure character similarity between the input photo and the generated 3D caricature, penalizing the identity change. As the input and output domains are rather different, we first reconstruct the 3D head of the input photo using the method introduced in Section~\ref{subsec:reconstruction_full_head}. 
Based on the cosine measure, $L_{\text{cha}}$ is defined as:
\begin{equation}
L_{\text{cha}}(G, \mathcal{P})
= \mathbb{E}_{p \sim \mathcal{P}}[1 - \overline{d_G}\cdot\overline{d_P}],
\label{eq:cha}
\end{equation}
where $d_G = v(G(p))-H_{\text{mean}}$ and $d_P = H(p)-H_{\text{mean}}$ are the feature vectors of $G(p)$ and $H(p)$, respectively\ylnew{.} $v(G(p))$ is the vector containing the vertices' coordinates of the mesh represented by caricature PCA parameter $G(p)$, \roberto{\ylnew{where} $\bar{v}$} is the normalized vector of $v$ and $\cdot$ is the dot product. 

{\it Caricature loss.} The cosine of the angle between the two feature vectors measures the face identity consistency of two faces, while the length of a feature vector $v$ measures the caricature style, i.e., the larger the magnitude $\|v\|$ ($\|v\|>1$), the more exaggerated the style is~\cite{hill2019deep}. Therefore, we propose a novel caricature loss $L_{\text{cari}}$ to constrain the caricature style:
\begin{equation}
L_{\text{cari}}(G, \mathcal{P})
= \mathbb{E}_{p \sim \mathcal{P}}\left[\exp\left(-\left(\overline{d_G}\cdot\overline{d_P}\right)\frac{\|d_G\|}{\|d_P\|}\right)\right],
\label{eq:cari}
\end{equation}
which penalizes identity inconsistency and insufficient exaggerations. 
The exponential form helps the magnitude to converge to a proper value: to decrease the loss term, $\|d_G\|$ will tend to be large, meanwhile since the exponential coefficient is negative, its gradient will decay exponentially. Furthermore, the adversarial $L_{\text{adv}}$ can constrain the magnitude of $d_G$ and make it converge to a proper value.

\begin{figure}[t]
  \centering
  \includegraphics[width=\columnwidth]{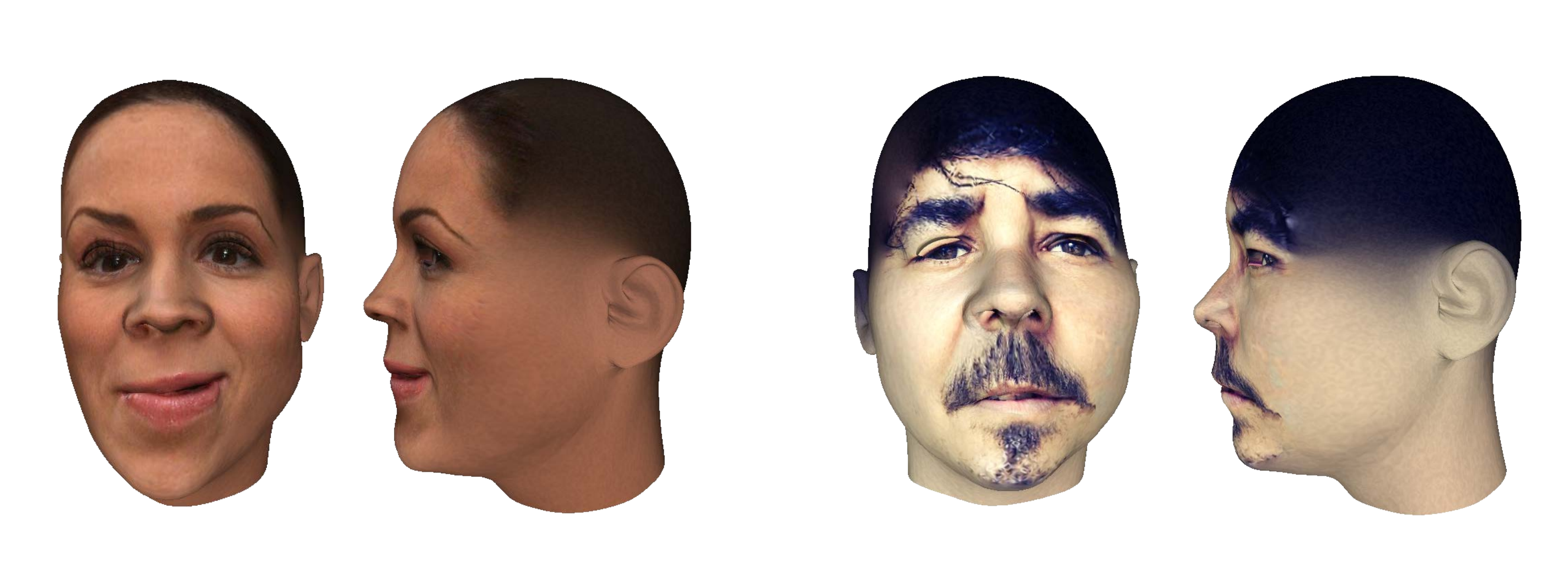}
  \vspace{-0.45in}
  \caption{
  The second option provided by our pipeline, which assigns color to each vertex of the full 3D caricature mesh.
  }
  \label{fig:fullhead_texture}
  \vspace{-0.1in}
\end{figure}

\textbf{Overall loss function.} The overall loss function for \ourGAN{} is:
\begin{equation*}
L_{\text{obj}} = L_{\text{adv}} + \lambda_{\text{cha}}L_{\text{cha}} + \lambda_{\text{cari}}L_{\text{cari}},
\end{equation*}
where $\lambda_{\text{cha}}$ and $\lambda_{\text{cari}}$ are the weights for balancing the multiple objectives. For all experiments, we set $\lambda_{\text{cha}} = 2$, $\lambda_{\text{cari}} = 20$.
We use the Adam solver to optimize the objective function for training the neural network.

\subsection{Caricature Texture Generation}
\label{subsec:texture}

Texture is essential for the appearance of 3D caricatures. In our pipeline, we provide two options so that a user can choose \roberto{the best for their application}. 

Based on the normal 3D head $H(p)$ reconstructed from input photo $p$, we detect common facial landmarks on both $H(p)$ and $p$. Then we compute a projection matrix according to the correspondence between these two sets of facial landmarks. This projection matrix enables us to build a texture mapping that maps the photo $p$ to the 3D head model $H(p)$. Since we ensure that the 3D caricature mesh $G(p)$ has the same connectivity as $H(p)$, there is a one-to-one correspondence between vertices of $G(p)$ and $H(p)$; therefore, the same texture mapping can be applied to $G(p)$. 
This texture \roberto{mapped} version of \roberto{the} 3D caricature is our first option. This option has a high texture resolution, but \roberto{the} texture only exists in \roberto{the (non-occluded) front of the face}. 

The first option does not have texture on the back of the head. In some applications such as 3D printing, the color on the \roberto{entire} head is needed. To 
achieve this, in our second option,
we store the color for each vertex \roberto{at the front of the face of the 3D} caricature (using the first option), and make up the color for remaining vertices. To do so, we use a smooth interpolation scheme which minimizes the Dirichlet energy by solving a linear system~\cite{ye2020dirichlet}. This interpolation is too smooth and lacks high-frequency texture information. To obtain a more natural texture, we calculate the variance of the front face texture and add random noise of the same variance to the remaining vertices. Two examples are shown in Figure \ref{fig:fullhead_texture}. This second option specifies the color at each vertex, \roberto{and is therefore of a lower} resolution than the first option; however, \roberto{colors are assigned over} the full head.

\subsection{Simple User Interaction}

\begin{figure}[htb]
  \centering
  \includegraphics[width=\columnwidth]{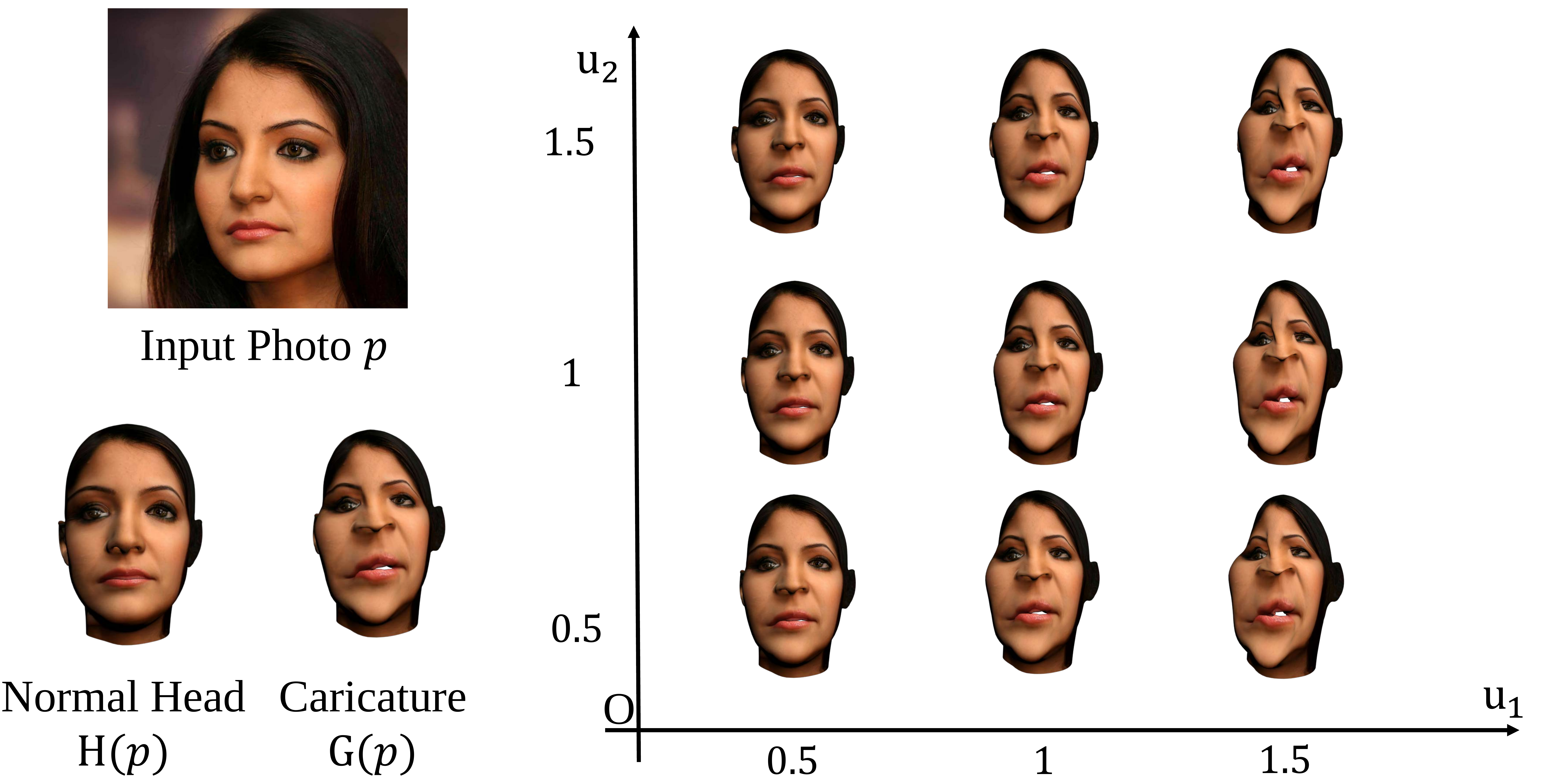}
  \vspace{-7mm}
  \caption{
  Examples of user control with different combinations of $u_1$ and $u_2$ in Eq.(\ref{eq:user-control}), i.e., $u_1 = 0.5, 1, 1.5$ and $u_2 = 0.5, 1, 1.5$.
  }\vspace{-0.2in}
  \label{fig:user_control}
\end{figure}

\begin{figure}[htb]
  \centering
  \includegraphics[width=\columnwidth]{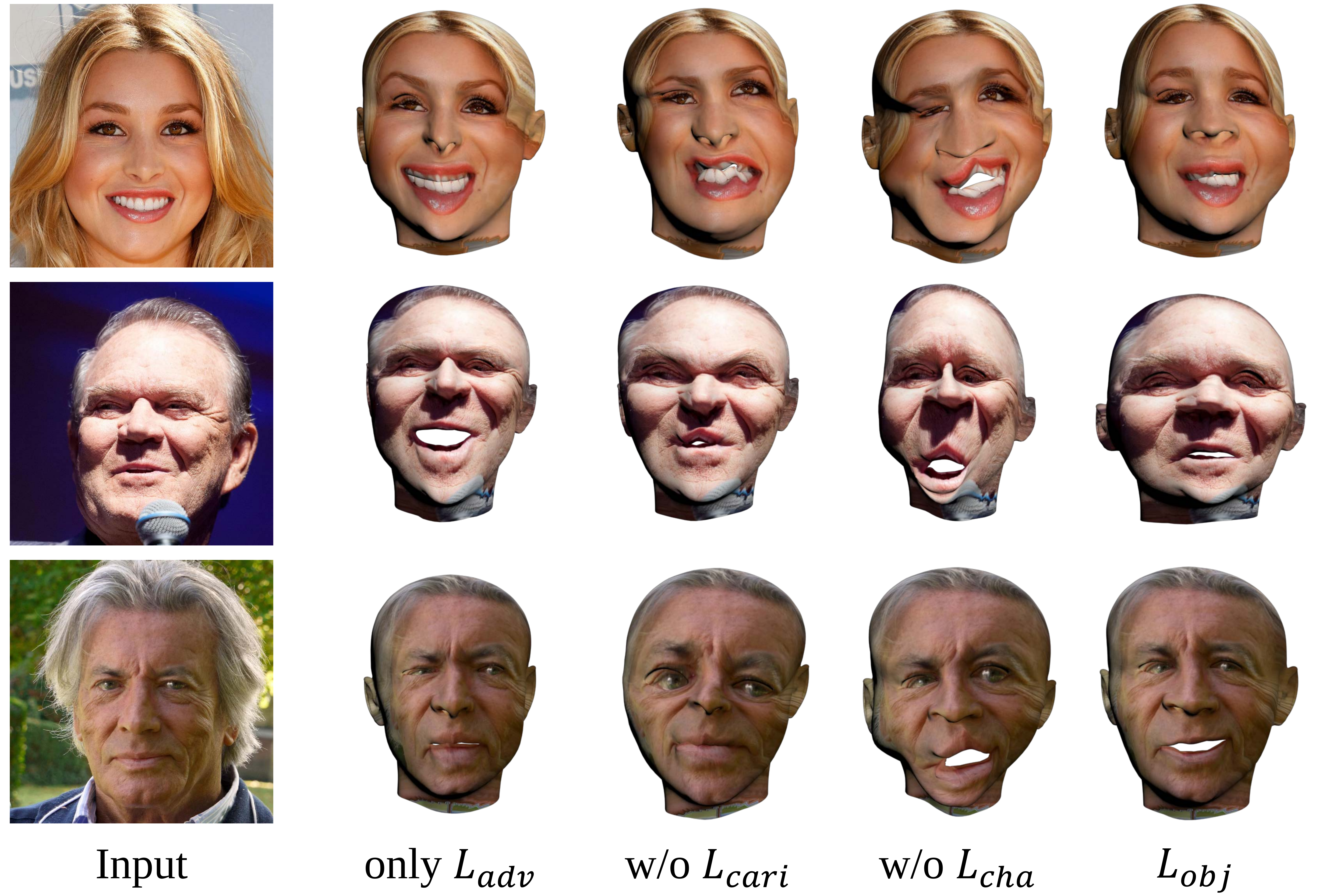}
  \vspace{-0.25in}
  \caption{
  Results for ablation study. The results with only $L_{\text{adv}}$ are highly random. The results without $L_{\text{cari}}$ are not exaggerated properly. 
  The results without $L_{\text{cha}}$ have low quality. Only whole method can generate ideal results.
  }\vspace{-0.05in}
  \label{fig:ablation_study}
\end{figure}

\begin{figure}[t]
\small
\centering
\includegraphics[width=\columnwidth]{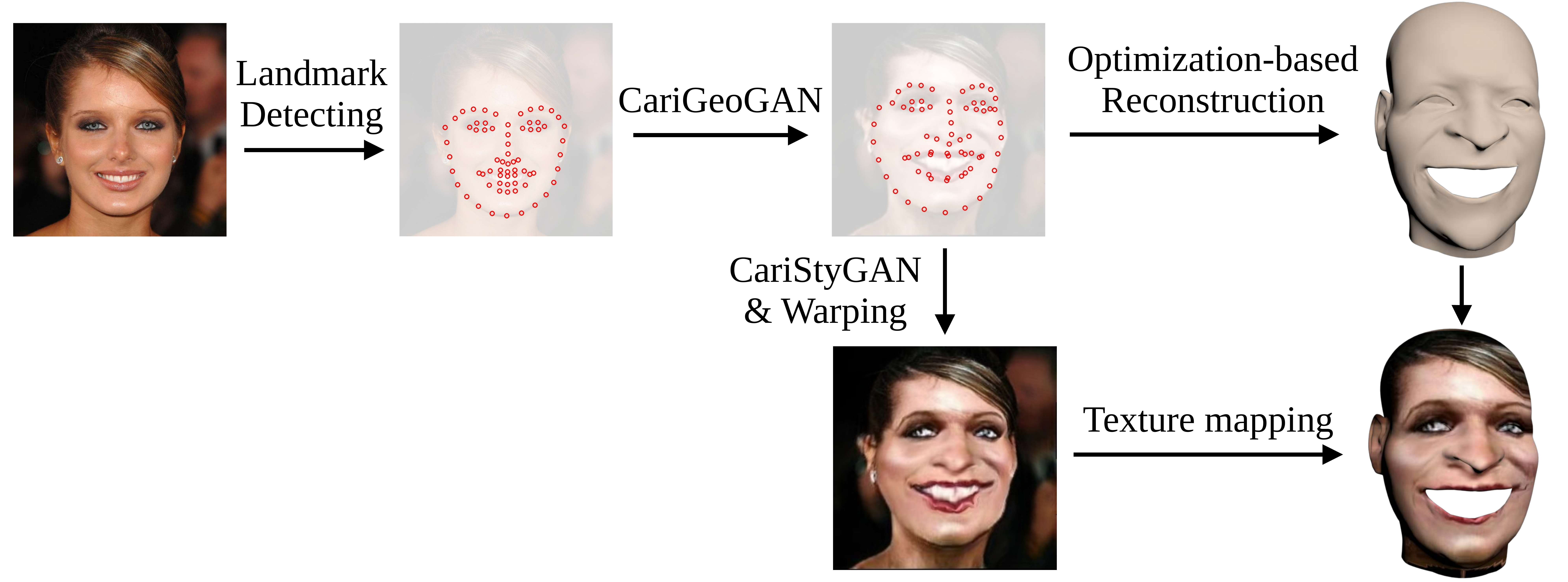}
\vspace{-0.15in}
\caption{The baseline method consists of several steps: landmark detection, landmark exaggeration using CariGeoGAN, image warping and generating 3D caricature meshes using facial landmarks.
}
\vspace{-0.1in}
\label{fig:baseline_method}
\end{figure}

A fully automatic method is convenient for users. However, sometimes they \roberto{may not be} satisfied
with the results \roberto{of} fully automatic methods. Therefore, allowing users to control the output in a simple way is desired. In our pipeline, given a 3D caricature $G(p)$ (automatically generated by 3D-CariGAN) and a normal 3D head model $H(p)$ (automatically reconstructed from input photo $p$), we extend the exaggeration equation in Eq.(\ref{eq:exaggeration}) to provide two simple-to-use and intuitive parameters (i.e., $u_1$ and $u_2$) for user control:
\begin{equation}
    C(p, u_1, u_2) = H_{\text{mean}} + u_1d_G +u_2d_P,
    \label{eq:user-control}
\end{equation}
where $d_G = v(G(p))-H_{\text{mean}}$ and $d_P = H(p)- H_{\text{mean}}$ are feature vectors of $G(p)$ and $H(p)$, respectively. 
Two parameters $u_1$ and $u_2$ have clear geometric meaning: (1) $u_2$ controls the exaggeration degree of $H(p)$ with the strict face identity; (2) $u_1$ controls the exaggeration degree of $G(p)$ with diverse exaggeration style; and (3) if $u_1+u_2=1$, the result is an interpolation between $G(p)$ and $H(p)$. An example of user control with different combinations of $u_1$ and $u_2$ is shown in Figure~\ref{fig:user_control}. 

\section{Experiments}

We have implemented the proposed \ourGAN{}, CariGeoGAN \cite{cao2018carigans} and normal 3D head reconstruction in PyTorch.
We have also implemented the optimization-based method~\cite{wu2018alive} in C++, as a baseline for comparison. We tested them on a PC with an Intel E5-2640v4 CPU (2.40 GHz) and an NVIDIA GeForce RTX 2080Ti GPU. The resolution of face photos in the CelebAMask-HQ dataset is $1024 \times 1024$. They are resized to $256 \times 256$ as input to \ourGAN{}.

In addition to the results presented in this section, the dataset details (Appendix A), evaluation of 3DCariPCA (Appendix B), implementation details (Appendix C) and more experimental results (Appendix D) are summarized in the appendix.

\subsection{Ablation Study}
\label{subsec:ablation}

We first show the benefits and necessity of using our PCA representation for 3D caricatures. As an alternative, we show the results of a variant of our method that instead uses the mesh vertex coordinates to represent 3D caricatures. Some results are shown in Figure~\ref{fig:noise_mesh}, which are very noisy and visually unacceptable, due to the higher dimensional space and lack of the facial constraint.

We then perform an ablation study to demonstrate the effectiveness of each loss term and some results are shown in Figure~\ref{fig:ablation_study}. We successively add adversarial loss, character loss and caricature loss to the objective function. The adversarial loss only ensures that the generation results are 3D caricatures, but the results are highly random. The character loss constrains the generation \roberto{results such that the identity and expression match} those of input photos, which make the generation results look like the input photos. The caricature loss strengthens the exaggeration style of the generation results. Character loss and caricature loss work well together. They ensure the generation results have good exaggeration styles and maintain good face identities. However, using only one of them cannot obtain good results.

\subsection{Comparison with Baseline Method}

To the best of the authors' knowledge, there is no existing method for generating 3D caricatures directly from photos. We build a baseline method by concatenating three methods. So far the optimization-based method~\cite{wu2018alive} is the only method for generating 3D caricature meshes from 2D caricature images. However, it needs facial landmarks of caricature images as input, which are difficult to obtain automatically. CariGeoGAN~\cite{cao2018carigans} is a method that exaggerates the facial landmarks extracted from face photos. We can warp the photo with the guidance of exaggerated landmarks using differentiable spline interpolation~\cite{Cole2017synthesizing}. Therefore, one possible way for generating 3D caricature meshes is to use a sequence of steps as shown in Figure~\ref{fig:baseline_method}. Obviously, the baseline method cannot allow users to adjust generation results interactively, while our method offers a simple tool for users to interactively adjust the generation results. For fair comparison, we use the same texture for the baseline method and our method.

Some comparison results of the proposed 3D-CariGAN and the baseline are shown in Figure~\ref{fig:results_cmp}. More results are shown in the appendix. For the baseline method, the exaggerated 2D facial landmarks are not directly designed for 3D caricatures, so the results of \roberto{the baseline method} can be too common or too strange. As a comparison, our method has a good exaggeration effect. In Section \ref{subsec:user-study}, we further conduct a user study for comparing the baseline and our method, which further demonstrates the advantage of our method. We also summarize the running times of both methods in Table~\ref{tab:times}, demonstrating that our method is much faster than the baseline.

\subsection{User Study}
\label{subsec:user-study}

Caricature is a kind of artistic style and so far there is still a lack of \roberto{suitable} objective evaluation methods. Therefore we design a two-level user study to compare our method with other methods.
\begin{table}[t]
\caption{Running time comparison of baseline and our method, averaged over $100$ results. The input is a $256 \times 256$ face photo 
and the output mesh has 12,124 vertices and 24,092 triangles. The running times of \ourGAN{} \yl{on} both CPU and GPU platforms are presented.
}
\vspace{-2mm}
\centering
\begin{tabular}{c|c|c}
\hline
Methods                  & Step           & Time (s) \\ \hline
\multirow{4}{*}{Baseline} & Landmark Detection & 0.086     \\ \cline{2-3}
& Warping          & 7.431     \\ \cline{2-3}
& CariGeoGAN           & 0.001     \\ \cline{2-3}
& Optimization-based reconstruction         & 14.510    \\ \hline
\multirow{2}{*}{\ourGAN{}}                     & Face Reconstruction (CPU)   & 1.109     \\ \cline{2-3}
                    & Other parts (CPU)  & 0.091     \\
\hline
\multirow{2}{*}{\ourGAN{}}                     & Face Reconstruction (GPU)            & 0.215     \\ \cline{2-3}
                    & Other parts (GPU)  & 0.011     \\
\hline
\end{tabular}
\label{tab:times}
\vspace{-5mm}
\end{table}

First we searched the Internet for caricature-related characteristics, and collected eleven terms: Uniform style, Reasonable structure, Clear theme, Color richness, Cultural connotation, Art skill, Creativeness, Similarity, Distinctiveness, Weirdness/Grotesque, and Exaggeration. We invited seven artists and asked them to select 3-5 \roberto{of the most} important characteristics for 3D caricatures. The selection results showed that the following four characteristics --- Reasonable structure, Similarity, Distinctiveness and Weirdness --- received more than half of the votes, and they characterized different aspects of 3D caricature. Then in the second-level user study, we invited participants to evaluate them by presenting the following criteria and explanations:

\begin{figure}[htb]
  \centering
  \includegraphics[width=0.5\textwidth]{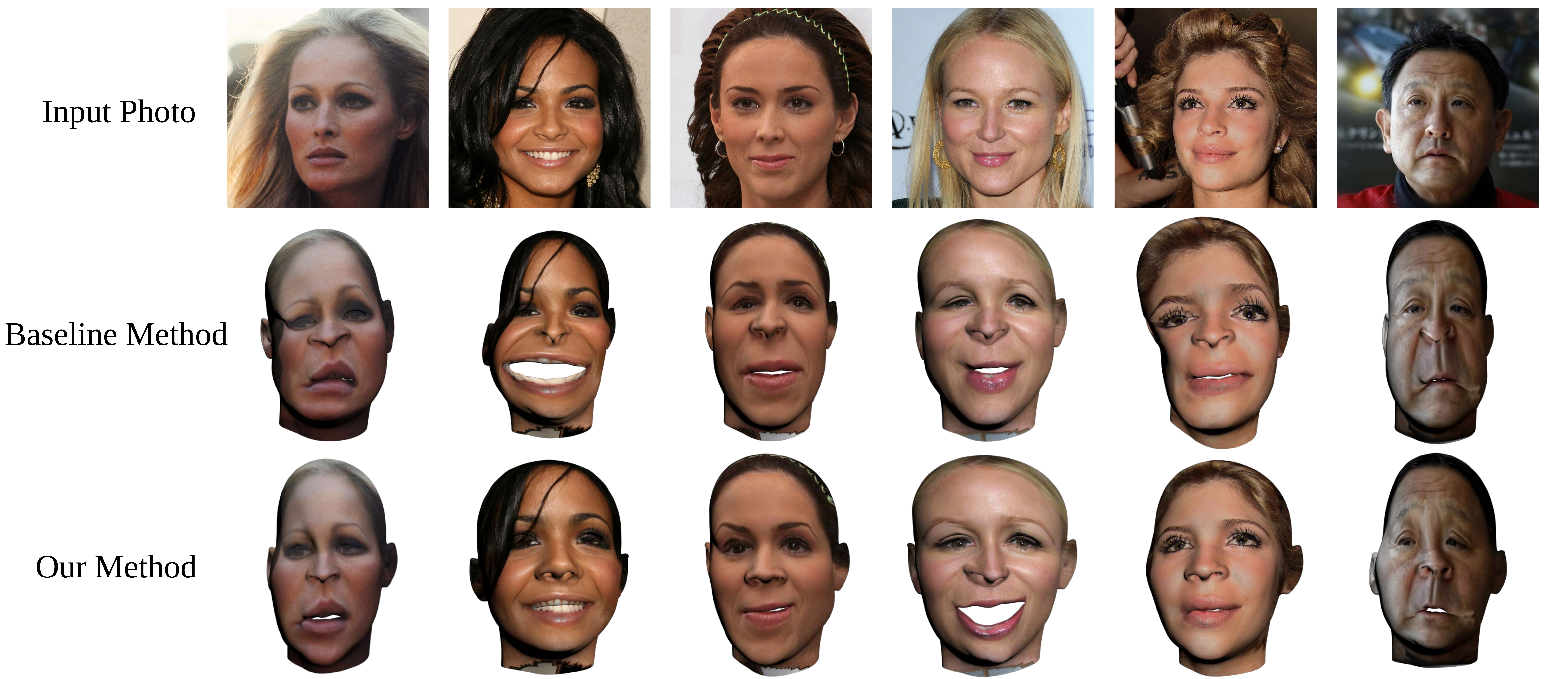}
  \vspace{-0.25in}
  \caption{Visual comparison between our method and baseline. Our user study in Section \ref{subsec:user-study} shows that our method has the best performance on RS, Sim and Dist, while baseline has the best performance on Weirdness. 
  }
  \label{fig:results_cmp}
  \vspace{-0.1in}
\end{figure}

\begin{itemize}
    \item \textbf{Reasonable Structure (RS).} The 3D caricature has reasonable structure looking like a human head. Please select the one which is most similar to a human head's structure.
    \item \textbf{Similarity (Sim).} The 3D caricature has the same or similar identity to the input face photo. Please select the one which is most similar to the face photo.
    \item \textbf{Distinctiveness (Dist).} The 3D caricature highlights the most significant part in the input face photo, instead of random deformation. Please select the one that catches the most significant characteristics of the face photo.
    \item \textbf{Weirdness.} The 3D caricature has weird and eerie aesthetic feeling. Please select the one which is most weird but makes you feel novel and beautiful.
\end{itemize}

\begin{table}[]
\caption{The average scores of four caricature characteristics in user study. The larger the score, the better the method is. 
}
\vspace{-2mm}
\centering
\begin{tabular}{c|c|c|c|c}
\hline
Method       & RS $\uparrow$    & Sim $\uparrow$ & Dist $\uparrow$ & Weirdness $\uparrow$ \\ \hline
Baseline     & 0.20  & 0.23       & 0.34            & \textbf{0.53}      \\ \hline
only $L_{\text{adv}}$ & 0.80  & 0.69       & 0.43            & 0.44      \\ \hline
w/o $L_{\text{cari}}$        & -0.28 & -0.29      & -0.03           & 0.11      \\ \hline
w/o $L_{\text{cha}}$        & -2.39 & -2.38      & -2.19           & -1.49     \\ \hline
Ours         & \textbf{1.66}  & \textbf{1.75}       & \textbf{1.45}            & 0.41      \\ \hline
\end{tabular}
\vspace{-5mm}
\label{table:votes}
\end{table}

Five methods were compared: baseline, three methods in the ablation study (only $L_{\text{adv}}$, w/o $L_{\text{cari}}$ and w/o $L_{\text{cha}}$) and our method 3D-CariGAN. 20 participants were 
recruited to conduct this user study. 10 face photos were used, and each photo has five 3D caricatures corresponding to five methods. For each photo, we randomly selected 5 pairs of 3D caricatures --- e.g., $(A_p,B_p)$, $(A_p,C_p)$, $(A_p,D_p)$, $(A_p,E_p)$ and $(B_p,C_p)$ from five caricatures $\{A_p,B_p,C_p,D_p,E_p\}$ corresponding to the photo $p$ --- and presented each pair to 10 participants. Finally 100 pairs of 3D caricatures were used and each participant watched 50 pairs. For each pair, the better one was selected for each of the four characteristics. Then 
we \roberto{convert the votes into}  a score using the ranking equation~\cite{chen2017learning}:
\begin{equation}
    score_p(I_i) = \sum_{I_j \in \Omega_p \setminus I_i} \frac{s_i - s_j}{s_{\text{max}}},
\label{eq:score}
\end{equation}
where $\Omega_p=\{A_p,B_p,C_p,D_p,E_p\}=\{I_i\}_{i=1}^5$ is a set of 5 caricature results for each photo $p$, $s_i$ is the number of votes for the result $I_i\in\Omega_p$, and $s_{\text{max}}=40$ is the maximum votes that each result can get. The range of score in Eq.(\ref{eq:score}) is $[-4,4]$. The larger the score, the better quality the 3D caricature has. Finally for each characteristic, the scores were averaged over 10 photos. The averaged scores for four characteristics \roberto{are} summarized in Table~\ref{table:votes}. These results show that our method \roberto{achieves} the best quality on RS, Sim and Dist, while the baseline has the best quality on Weirdness. A possible reason is that the baseline explicitly uses 2D caricature information to guide the generation of 3D caricature, without imposing the constraint on face identity to the input photo, so it can generate a better weirdness effect but \roberto{is worse on the other} three characteristics. As a comparison, our method achieves a good balance on \roberto{all} four characteristics.

\section{Discussions \& Conclusions}

{\it Identity and Expression Preservation Ability.} We design a quantitative evaluation for identity and expression preservation ability of our method. The details and results are presented in the appendix.

{\it Limitations.} In this paper, we only generate exaggerated geometry and directly use the texture of a photo. It is \roberto{difficult} to generate texture for a 3D caricature because 2D caricature is \roberto{not a suitable reference}. To address this challenge, 3D caricatures with textures created by artists are helpful. We \roberto{could} simultaneously model geometry and texture \roberto{using} neural networks if we had this kind of dataset. Another limitation is that the complete textures is defined as colors at mesh vertices. Generating complete texture in terms of texture mapping needs to be explored in the future.

In this paper, we propose an end-to-end deep neural network model that transforms a normal face photo into a 3D caricature, which is an extreme cross-domain task. To accomplish this task, we build a 3D caricature dataset, establish a PCA model, and propose two novel loss terms based on previous psychological studies. Our method is fast and thus makes further interaction control possible. We also propose a simple and intuitive method that allows a user to interactively adjust the results. Experiments and a user study demonstrate the effectiveness of our method.


%

\ifCLASSOPTIONcompsoc
\else
  \section*{Acknowledgment}
\fi


\ifCLASSOPTIONcaptionsoff
  \newpage
\fi

\bibliographystyle{IEEEtran}
\bibliography{main}

\begin{thebibliography}{10}
\providecommand{\url}[1]{#1}
\csname url@samestyle\endcsname
\providecommand{\newblock}{\relax}
\providecommand{\bibinfo}[2]{#2}
\providecommand{\BIBentrySTDinterwordspacing}{\spaceskip=0pt\relax}
\providecommand{\BIBentryALTinterwordstretchfactor}{4}
\providecommand{\BIBentryALTinterwordspacing}{\spaceskip=\fontdimen2\font plus
\BIBentryALTinterwordstretchfactor\fontdimen3\font minus
  \fontdimen4\font\relax}
\providecommand{\BIBforeignlanguage}[2]{{%
\expandafter\ifx\csname l@#1\endcsname\relax
\typeout{** WARNING: IEEEtran.bst: No hyphenation pattern has been}%
\typeout{** loaded for the language `#1'. Using the pattern for}%
\typeout{** the default language instead.}%
\else
\language=\csname l@#1\endcsname
\fi
#2}}
\providecommand{\BIBdecl}{\relax}
\BIBdecl

\bibitem{sadimon2010computer}
S.~B. Sadimon, M.~S. Sunar, D.~Mohamad, and H.~Haron, ``Computer generated
  caricature: A survey,'' in \emph{International Conference on Cyberworlds
  ({CW})}.\hskip 1em plus 0.5em minus 0.4em\relax IEEE, 2010, pp. 383--390.

\bibitem{brennan2007caricature}
S.~E. Brennan, ``Caricature generator: The dynamic exaggeration of faces by
  computer,'' \emph{Leonardo}, vol.~40, no.~4, pp. 392--400, 2007.

\bibitem{han2018caricatureshop}
X.~Han, K.~Hou, D.~Du, Y.~Qiu, S.~Cui, K.~Zhou, and Y.~Yu, ``{CaricatureShop}:
  Personalized and photorealistic caricature sketching,'' \emph{{IEEE} Trans.
  Vis. Comput. Graph.}, vol.~26, no.~7, pp. 2349--2361, 2020.

\bibitem{li2018carigan}
W.~Li, W.~Xiong, H.~Liao, J.~Huo, Y.~Gao, and J.~Luo, ``{CariGAN}: Caricature
  generation through weakly paired adversarial learning,''
  \emph{arXiv:1811.00445}, 2018.

\bibitem{cao2018carigans}
K.~Cao, J.~Liao, and L.~Yuan, ``{CariGANs}: Unpaired photo-to-caricature
  translation,'' \emph{ACM Trans. Graph.}, vol.~37, no.~6, pp. 244:1--244:14,
  Dec. 2018.

\bibitem{wu2018alive}
Q.~Wu, J.~Zhang, Y.-K. Lai, J.~Zheng, and J.~Cai, ``Alive caricature from {2D}
  to {3D},'' in \emph{IEEE CVPR}, 2018, pp. 7336--7345.

\bibitem{han2017deepsketch2face}
X.~Han, C.~Gao, and Y.~Yu, ``{DeepSketch2Face}: a deep learning based sketching
  system for {3D} face and caricature modeling,'' \emph{ACM Trans. Graph.},
  vol.~36, no.~4, p. 126, 2017.

\bibitem{Cole2017synthesizing}
F.~Cole, D.~Belanger, D.~Krishnan, A.~Sarna, I.~Mosseri, and W.~T. Freeman,
  ``Synthesizing normalized faces from facial identity features,'' in
  \emph{IEEE CVPR}, 2017.

\bibitem{blanz1999morphable}
V.~Blanz and T.~Vetter, ``A morphable model for the synthesis of {3D} faces,''
  in \emph{ACM SIGGRAPH}, 1999, pp. 187--194.

\bibitem{CelebAMask-HQ}
C.-H. Lee, Z.~Liu, L.~Wu, and P.~Luo, ``{MaskGAN}: Towards diverse and
  interactive facial image manipulation,'' \emph{arXiv:1907.11922}, 2019.

\bibitem{hill2019deep}
M.~Q. Hill, C.~J. Parde, C.~D. Castillo, Y.~I. Colon, R.~Ranjan, J.-C. Chen,
  V.~Blanz, and A.~J. O’Toole, ``Deep convolutional neural networks in the
  face of caricature,'' \emph{Nature Machine Intelligence}, vol.~1, no.~11, pp.
  522--529, 2019.

\bibitem{Benson1991}
P.~J. Benson and D.~I. Perrett, ``Perception and recognition of photographic
  quality facial caricatures: Implications for the recognition of natural
  images,'' \emph{European Journal of Cognitive Psychology}, vol.~3, no.~1, pp.
  105--135, 1991.

\bibitem{Rhodes1997}
G.~Rhodes, G.~Byatt, T.~Tremewan, and A.~Kennedy, ``Facial distinctiveness and
  the power of caricatures,'' \emph{Perception}, vol.~26, pp. 207--223, 1997.

\bibitem{gatys2015neural}
L.~A. Gatys, A.~S. Ecker, and M.~Bethge, ``A neural algorithm of artistic
  style,'' \emph{arXiv:1508.06576}, 2015.

\bibitem{liao2017visual}
J.~Liao, Y.~Yao, L.~Yuan, G.~Hua, and S.~B. Kang, ``Visual attribute transfer
  through deep image analogy,'' \emph{ACM Trans. Graph.}, vol.~36, no.~4, p.
  120, 2017.

\bibitem{shi2019warpgan}
Y.~Shi, D.~Deb, and A.~K. Jain, ``{WarpGAN}: Automatic caricature generation,''
  in \emph{IEEE CVPR}, 2019, pp. 10\,762--10\,771.

\bibitem{zollhofer2018state}
M.~Zollh{\"o}fer, J.~Thies, P.~Garrido, D.~Bradley, T.~Beeler, P.~P{\'e}rez,
  M.~Stamminger, M.~Nie{\ss}ner, and C.~Theobalt, ``State of the art on
  monocular {3D} face reconstruction, tracking, and applications,''
  \emph{Computer Graphics Forum}, vol.~37, no.~2, pp. 523--550, 2018.

\bibitem{paysan20093d}
P.~Paysan, R.~Knothe, B.~Amberg, S.~Romdhani, and T.~Vetter, ``A {3D} face
  model for pose and illumination invariant face recognition,'' in \emph{IEEE
  International Conference on Advanced Video and Signal Based Surveillance
  ({AVSS})}, 2009, pp. 296--301.

\bibitem{booth2018large}
J.~Booth, A.~Roussos, A.~Ponniah, D.~Dunaway, and S.~Zafeiriou, ``Large scale
  {3D} morphable models,'' \emph{Intl. J. Comp. Vis.}, vol. 126, no. 2-4, pp.
  233--254, 2018.

\bibitem{dai20173d}
H.~Dai, N.~Pears, W.~A. Smith, and C.~Duncan, ``A {3D} morphable model of
  craniofacial shape and texture variation,'' in \emph{IEEE ICCV}, 2017, pp.
  3085--3093.

\bibitem{vlasic2005face}
D.~Vlasic, M.~Brand, H.~Pfister, and J.~Popovi{\'c}, ``Face transfer with
  multilinear models,'' \emph{ACM Trans. Graph.}, vol.~24, no.~3, pp. 426--433,
  2005.

\bibitem{cao2013facewarehouse}
C.~Cao, Y.~Weng, S.~Zhou, Y.~Tong, and K.~Zhou, ``{FaceWarehouse}: A {3D}
  facial expression database for visual computing,'' \emph{IEEE Trans. Vis.
  Comp. Graph.}, vol.~20, no.~3, pp. 413--425, 2013.

\bibitem{jackson2017large}
A.~S. Jackson, A.~Bulat, V.~Argyriou, and G.~Tzimiropoulos, ``Large pose {3D}
  face reconstruction from a single image via direct volumetric {CNN}
  regression,'' in \emph{IEEE ICCV}, 2017, pp. 1031--1039.

\bibitem{tewari2017mofa}
A.~Tewari, M.~Zollhofer, H.~Kim, P.~Garrido, F.~Bernard, P.~Perez, and
  C.~Theobalt, ``Mofa: Model-based deep convolutional face autoencoder for
  unsupervised monocular reconstruction,'' in \emph{IEEE ICCV}, 2017, pp.
  1274--1283.

\bibitem{jiang20183d}
L.~Jiang, J.~Zhang, B.~Deng, H.~Li, and L.~Liu, ``{3D} face reconstruction with
  geometry details from a single image,'' \emph{IEEE Trans. Image Processing},
  vol.~27, no.~10, pp. 4756--4770, 2018.

\bibitem{sela2015computational}
M.~Sela, Y.~Aflalo, and R.~Kimmel, ``Computational caricaturization of
  surfaces,'' \emph{Computer Vision and Image Understanding}, vol. 141, pp.
  1--17, 2015.

\bibitem{clarke2010automatic}
L.~Clarke, M.~Chen, and B.~Mora, ``Automatic generation of {3D} caricatures
  based on artistic deformation styles,'' \emph{IEEE Trans. Vis. Comp. Graph.},
  vol.~17, no.~6, pp. 808--821, 2010.

\bibitem{ranjan2018generating}
A.~Ranjan, T.~Bolkart, S.~Sanyal, and M.~J. Black, ``Generating {3D} faces
  using convolutional mesh autoencoders,'' in \emph{ECCV}, 2018, pp. 704--720.

\bibitem{Huo:2017:VRC:3126686.3126736}
J.~Huo, Y.~Gao, Y.~Shi, and H.~Yin, ``Variation robust cross-modal metric
  learning for caricature recognition,'' in \emph{ACM Multimedia Conference},
  2017, pp. 340--348.

\bibitem{HuoBMVC2018WebCaricature}
J.~Huo, W.~Li, Y.~Shi, Y.~Gao, and H.~Yin, ``{WebCaricature}: a benchmark for
  caricature recognition,'' in \emph{British Machine Vision Conference
  ({BMVC})}, 2018.

\bibitem{king2009dlib}
D.~E. King, ``Dlib-ml: A machine learning toolkit,'' \emph{Journal of Machine
  Learning Research}, vol.~10, no. Jul, pp. 1755--1758, 2009.

\bibitem{deng2019accurate}
Y.~Deng, J.~Yang, S.~Xu, D.~Chen, Y.~Jia, and X.~Tong, ``Accurate {3D} face
  reconstruction with weakly-supervised learning: From single image to image
  set,'' in \emph{IEEE CVPR Workshops}, 2019.

\bibitem{amberg2007}
B.~{Amberg}, S.~{Romdhani}, and T.~{Vetter}, ``Optimal step nonrigid {ICP}
  algorithms for surface registration,'' in \emph{IEEE Conference on Computer
  Vision and Pattern Recognition}, 2007, pp. 1--8.

\bibitem{ploumpis2019combining}
S.~Ploumpis, H.~Wang, N.~Pears, W.~A. Smith, and S.~Zafeiriou, ``Combining {3D}
  morphable models: A large scale face-and-head model,'' in \emph{IEEE CVPR},
  2019, pp. 10\,934--10\,943.

\bibitem{he2016deep}
K.~He, X.~Zhang, S.~Ren, and J.~Sun, ``Deep residual learning for image
  recognition,'' in \emph{IEEE CVPR}, 2016, pp. 770--778.

\bibitem{mao2017least}
X.~Mao, Q.~Li, H.~Xie, R.~Y. Lau, Z.~Wang, and S.~Paul~Smolley, ``Least squares
  generative adversarial networks,'' in \emph{IEEE ICCV}, 2017, pp. 2794--2802.

\bibitem{ye2020dirichlet}
Z.~Ye, R.~Yi, W.~Gong, Y.~He, and Y.-J. Liu, ``Dirichlet energy of {Delaunay}
  meshes and intrinsic {Delaunay} triangulations,'' \emph{Computer-Aided
  Design}, vol. 126, p. 102851, 2020.

\bibitem{chen2017learning}
Y.~Chen, Y.-J. Liu, and Y.-K. Lai, ``Learning to rank retargeted images,'' in
  \emph{IEEE CVPR}, 2017, pp. 3994--4002.

\end{thebibliography}

\end{document}